\documentclass[runningheads]{llncs}
\usepackage{xcolor}
\usepackage[algoruled,boxed,lined]{algorithm2e}
\SetKwInput{kwhyper}{Hyperparameters}
\usepackage{algcompatible}
\usepackage{graphicx}  
\usepackage{subfig}
\SetKwRepeat{Do}{do}{while}%
\usepackage{float}
\usepackage{hyperref}
\usepackage{todonotes}
\usepackage{amsmath,amssymb}
\usepackage{multirow}

\newcommand{\action}{a}
\newcommand{\signal}{I}
\newcommand{\reward}{r}
\newcommand{\transition}{T}
\newcommand{\state}{s}
\newcommand{\Qhat}[2]{\hat{Q}_{#1}(#2)}

\newcommand{\model}{M}
\newcommand{\Qtree}[2]{Q^{UT}_{#1}(#2)}
\DeclareMathOperator{\E}{\mathbb{E}}

\title{Toward Interpretable Deep Reinforcement Learning with Linear Model U-Trees}

\begin{document}
\author{Guiliang Liu, Oliver Schulte, Wang Zhu and Qingcan Li}
\institute{School of Computing Science, Simon Fraser University,\\\email{gla68@sfu.ca, oschulte@cs.sfu.ca, zhuwangz@sfu.ca, qingcanl@sfu.ca}}

\titlerunning{Toward Interpretable DRL with Linear Model U-Trees}
\authorrunning{Guiliang Liu, Oliver Schulte, Wang Zhu and Qingcan Li}
\maketitle

\begin{abstract}
    
    
    Deep Reinforcement Learning (DRL) has achieved impressive success in many applications. A key component of many DRL models is a neural network representing a Q function, to estimate the expected cumulative reward following a state-action pair. The Q function neural network contains a lot of implicit knowledge about the RL problems, but often remains unexamined and uninterpreted. To our knowledge, this work develops the first mimic learning framework for Q functions in DRL.
    We introduce Linear Model U-trees (LMUTs) to approximate neural network predictions. An LMUT is learned using a novel on-line algorithm that is well-suited for an active play setting, where the mimic learner observes an ongoing interaction between the neural net and the environment. Empirical evaluation shows that an LMUT mimics a Q function 
    substantially better than five baseline methods. The transparent tree structure of an LMUT facilitates understanding the network's learned knowledge by analyzing feature influence, extracting rules, and highlighting the super-pixels in image inputs.


    
\end{abstract}

\section{Introduction: Mimic a Deep Reinforcement Learner}

Deep Reinforcement Learning has mastered
human-level control policies in a wide variety of tasks~\cite{Mnih2015}.
Despite excellent performance, the learned knowledge remains implicit in neural networks and hard to explain.
There exists a trade-off between model performance and interpretability~\cite{lipton2016mythos}.
%
One of the methods to address this trade-off is mimic learning~\cite{ba2014deep}, which 
trains an interpretable mimic model to match the predictions of a highly accurate model. Many works~\cite{che2016interpretable,boz2002extracting,dancey2007logistic} have applied types of mimic learning 
to distill knowledge from deep models to a mimic model with tree representation. 
Current methods focus only on interpreting deep models for supervised learning.
However, DRL is an unsupervised process, where agents continuously interact with an environment, instead of learning from a static training/testing dataset. 

This work develops a novel mimic learning framework for Reinforcement Learning.
We examine two different approaches to {\em generating data for RL mimic learning}.
Within the first {\em Experience Training} setting, which allows applying traditional batch learning methods to train a mimic model, we record all state action pairs during the training process of DRL and complement them with Q values as soft supervision labels. 
Storing and reading the training experience of a DRL model consumes much time and space, and the training experience may not even be available to a mimic learner. 
%
Therefore our second {\em Active Play} setting  
generates streaming data through interacting with the environment using the mature DRL model. 
The active play setting requires an on-line algorithm to dynamically 
update the model as more learning data is generated.

U-tree~\cite{mccallum1996learning,uther1998tree} is a classic online reinforcement learning method which represents a Q function using a tree structure.
To strengthen its generalization ability, 
we add a linear model to each leaf node, which defines a novel Linear Model U-Tree (LMUT).
To support the active play setting, we introduce a novel on-line learning algorithm for LMUT, which applies Stochastic Gradient Descent to update the linear models, 
given some memory of recent input data stored on each leaf node.  
We conducted an empirical evaluation in three benchmark environments with five baseline methods.  Two natural evaluation metrics for an RL mimic learner are: 1) fidelity 
\cite{dancey2007logistic}: how well the mimic model matches the predictions of the neural net, as in supervised learning, and 2) {\em play performance}: how well the average return achieved by a controller based on the mimic model matches the return achieved by the neural net. Play performance is the most relevant metric for  reinforcement learning. Perfect fidelity implies a perfect match in play performance. However, our experiments show that approximate fidelity does not imply a good match in play performance. This is because RL mimic learning must strike a balance between coverage: matching the neural net across a large section of the state space, and optimality: matching the neural net on the states that are most important for performance.
In our experiments, LMUT learning achieves a good balance: the best match to play performance among the mimic methods, and competitive fidelity to the neural net predictions. 
The transparent tree structure of LMUT makes the DRL neural net interpretable.
To analyze the mimicked knowledge, we calculate the importance of input features and extract rules for typical examples of agent behavior. 
 For image inputs, the super-pixels in input images are highlighted to illustrate the key regions.\\

\noindent\textbf{Contributions.}
The main contributions of this paper are  as follow:
1) To our best knowledge, the first work that extends interpretable mimic learning to Reinforcement Learning.
2) A novel on-line learning algorithm for LMUT, a novel model tree to mimic a DRL model.
3) We show how to interpret a DRL model by analyzing the knowledge stored in the 
tree structure of LMUT.\\

The paper is organized as follow. Section 2 covers the background and related work of DRL, mimic learning and U-tree. Section 3 introduces the mimic learning framework and Section 4 shows how to learn a LMUT. Empirical evaluation is performed in section 5 and section 6 discusses the interpretability of LMUT.




\section{Background and Related Work}
{\em Reinforcement Learning and the Q-function.}
Reinforcement Learning constructs a policy for agents to interact with environment and maximize cumulative reward 
~\cite{sutton1998introduction}. Such an environment can be formalized as a Markov Decision Process (MDP) with 4-tuple $(S,A,P,R)$, where at timestep t, an agent observes a state $\state_{t}\in S$, chooses a action $\action_{t}\in A$ and receives a reward $\reward_{t}\in R$ and the next observation $s_{t+1}\in S$ from environment.
A Q function represents the value of executing action $\action_{t}$ under state $\state_{t}$ \cite{riedmiller2005neural}. Given a policy $\pi$, the value is the expectation of the sum of discounted reward $Q_{t}(\state_{t},\action_{t}) = \E_{\pi}(\sum_{k=0}^{\infty}\gamma^{k} r_{t+k+1})$. $Q$-learning is similar to temporal difference methods that update the current Q value estimates towards the observed reward and estimated utility of the resulting state $\state_{t+1}$. 
An advanced model Deep Q-Network (DQN)~\cite{Mnih2015} was proposed, which uses neural network to approximate the $Q$ function approximation.
Parameter ($\theta$) updates 
minimize the differentiable loss function: 
\begin{align}
\mathcal{L}(\theta_{i}) &\approx (\reward_{t}+\gamma \max_{\action_{t}} Q(\state_{t+1},\action_{t+1}|\theta_{i}) - Q(\state_{t},\action_{t}|\theta_{i}))^{2}\\
    \theta_{i+1} & = \theta_{i}+\alpha \nabla_{\theta}\mathcal{L}(\theta_{i})
\end{align}

{\em Mimic Learning.}
Recent works on mimic learning ~\cite{ba2014deep,che2016interpretable,dancey2007logistic} have demonstrated that models like shallow feed-forward neural network or decision trees can mimic the function of a deep neural net with complex structures. In the {\em oracle framework}, soft output labels are collected by passing inputs to 
a large, complex and accurate deep neural network.
Then we train a mimic model with the soft output as supervisor.
The results indicate that training a mimic model with soft output achieves substantial improvement in accuracy and efficiency, over training the same model type directly with hard targets from the dataset. 
But previous works studied only supervised learning (classification/prediction), rather than Reinforcement Learning as in our work. 

{\em U-Tree Learning.}
A tree structure is transparent and interpretable, allowing rule extraction and measuring feature influence~\cite{che2016interpretable}.
U-tree~\cite{mccallum1996learning} learning was developed as an online reinforcement learning algorithm with a tree structure representation.
A U-tree takes a set of observed feature/action values as input and maps it to a state value (or Q-value).
~\cite{uther1998tree} introduces the continuous U-tree (CUT) for 
continuous state features.
CUT learning dynamically generates a tree-based discretization of the input signal and estimates state transition probabilities 
by retaining transitions in every leaf node~\cite{uther1998tree}.
 CUT learning applies dynamic programming to solve it to solve the resulting Markov Decision Process (MDP).
Although CUT has been successfully applied in test environments like Corridor and Hexagonal Soccer, constructing Continuous U-tree from raw data is rather slow and consumes much computing time and space.

\section{Mimic Learning  for Deep Reinforcement Learning}
Unlike supervised learning, a DRL model 
is not trained with static input/output data pairs; instead it interacts with the environment by selecting actions to perform and  adjusting its policy to maximize the expectation of cumulative reward.
We now present two settings to mimic the Q functions in DRL models.

\subsubsection{Experience Training}
generates data for batch training, following~\cite{ba2014deep,che2016interpretable}.
To construct a mimic 
dataset, we record all the observation signals $\signal$ and actions $\action$ during the DRL process. A signal $\signal$ is a vector of continuous features that represents a state (for discrete features we use one-hot representation). Then, by inputting them to a mature DRL model, we obtain their corresponding soft output $Q$ and use the entire input/output pairs $\{(\langle\signal_{1},\action_{1}\rangle,\Qhat{1}{\signal_{1},\action_{1}}),(\langle\signal_{2},\action_{2}\rangle,\Qhat{2}{\signal_{2},\action_{2}})$ $,...,(\langle\signal_{T},\action_{T}\rangle,\Qhat{T}{\signal_{T},\action_{T}})\}$ as the \textbf{experience training dataset}.
. 

\begin{figure}[htbp]
\centering
\includegraphics[width=0.9\textwidth] 
{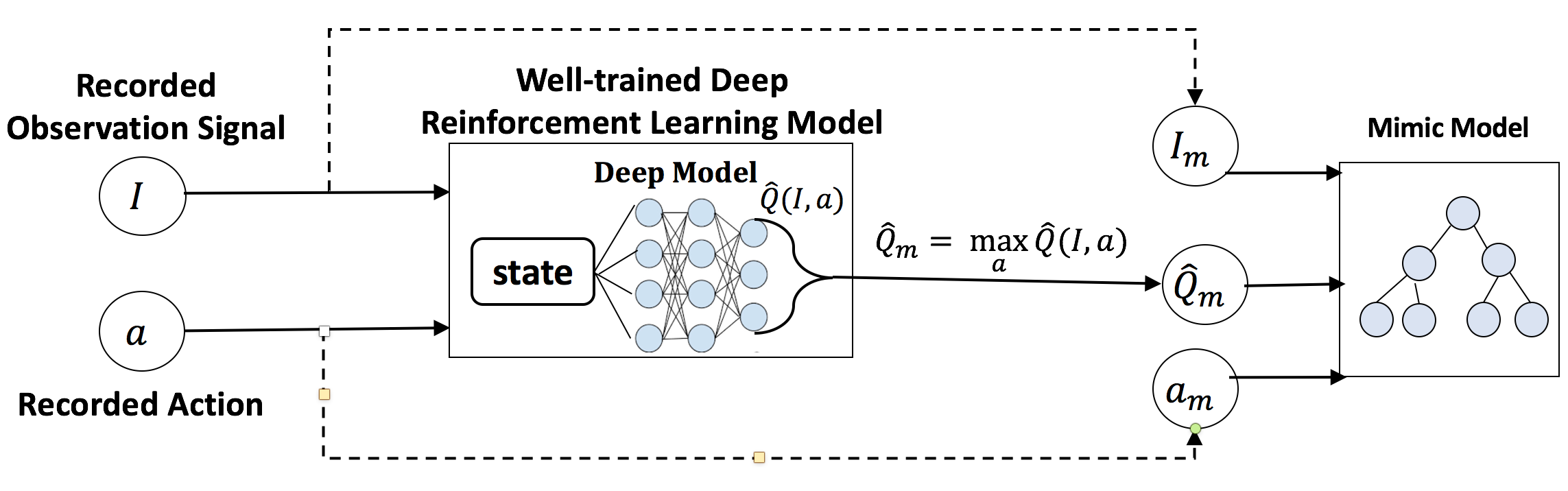}
\caption{Experience Training Setting
}
\label{fig:experience-training-setting}
\end{figure}


\subsubsection{Active Play} 
generates mimic data by applying a mature DRL model to interact with the environment.
Similar to
~\cite{tong2001support}, our active learner $\ell$ has three components: $(q,f,I)$. The first component $q$ is a querying function $q(I)$ that gives the current observed signal $\signal$, selects an action $\action$.
The querying function controls $\ell$'s interaction with the environment so it must consider the balance between exploration and exploitation. Here the $\epsilon$-greedy scheme~\cite{Mnih2015} ($\epsilon$ decaying from 1 to 0) is used as our querying function. 
The second component $f$ is the deep model that produces Q values: $f:(I, a)\rightarrow \it{range}(\hat{Q})$. 

\begin{figure}[htbp]
\centering
\includegraphics[width=0.9\textwidth] 
{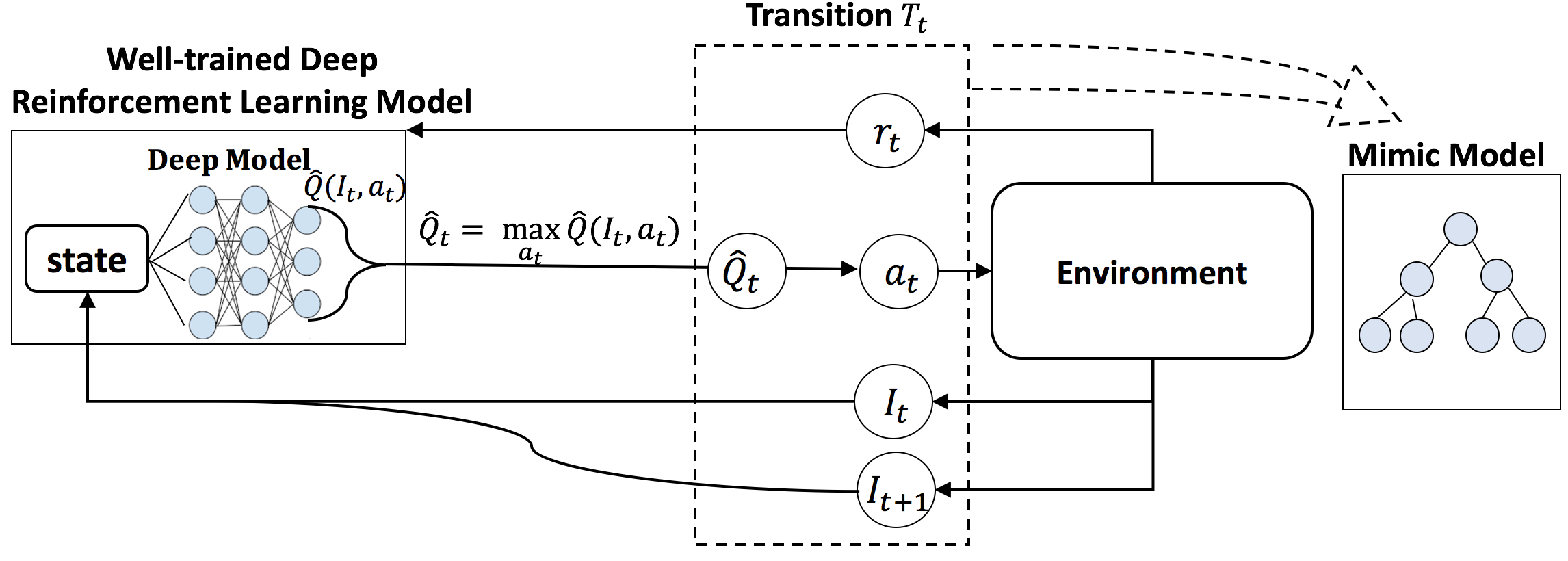}
\caption{Active Play Setting.
}
\label{fig:active-play-setting}
\end{figure}

As shown in Figure~\ref{fig:active-play-setting}, the mimic training data is generated in the following steps:
{\it Step 1:} Given a starting observation signal $\signal_{t}$ on time step $t$, we select an action $\action_{t}=q(\signal_{t})$, 
and obtain a soft output Q value $\hat{Q}_{t} =f(\signal_{t}, \action_{t}) $.
{\it Step 2:} After performing $\action_{t}$, 
the environment 
provides a reward $\reward_{t}$ and the next state observation $\signal_{t+1}$ 
.
We record a labelled \textbf{transition} $\transition_{t}=\{\signal_{t}, \action_{t}, \reward_{t}, \signal_{t+1}, \hat{Q}_{t}(\signal_{t},\action_{t}) \}$ where the soft label $ \hat{Q}_{t}(\signal_{t},\action_{t})$ comes from the well trained DRL model. 
{\it Step 3:} We set $\signal_{t+1}$ as the next starting observation signal, repeat above steps until we have training data for the active learner $\ell$ to finish sufficient updates over mimic model $m$. This process will produce an infinite data stream (transitions $\transition$) in sequential order. We use minibatch online learning, where the learner returns a mimic model $\model$ after some fixed batchsize $B$ of queries.

Compared to Experience Training, Active Play does not require recording data during the training process of DRL models. This is important because: (1) Many mimic learners have access only to the trained deep models. (2) Training a DRL model often generates a large amount of data, 
which requires much memory and is computationally challenging to process.
(3) The Experience Training data includes frequent visits to suboptimal states, 
which makes it difficult for the mimic learner to obtain an optimal return. 


\section{Learning Linear Model U-Trees}
A neural network with continuous activation functions computes a continuous function. A regression tree can approximate a continuous function arbitrarily closely, given enough leaves. Continuous U-Trees (CUTs) are essentially regression trees for value functions, and therefore a natural choice for a tree structure representation of a DRL Q function. However, their ability to generalize is limited, and CUT learning converges slowly.  In this paper, we introduce a novel extension of CUT, Linear Model U-Tree (LMUT), that allows CUT leaf nodes to contain a linear model, rather than simple constants. Being strictly more expressive than a regression tree, a linear model tree can also approximate a continuous function arbitrarily closely, with typically with many fewer leaves~\cite{chaudhuri1994piecewise}. Smaller trees are more interpretable, and therefore more suitable for mimic learning.

As shown in Figure~\ref{fig:u-tree-structure} and Table~\ref{table:partition-cell}, each leaf node
of a LMUT defines a partition cell of the input space, which can be interpreted as a discrete state $s$ for the decision process.
Within each partition cell, LMUT also records the reward $\reward$ and the transition probabilities $p$ of performing action $a$ on the current state $s$, as shown in the Leaf Node 5 of Figure~\ref{fig:u-tree-structure}.
So LMUT builds a Markov Decision Process (MDP) from the interaction data between environment and deep model.
Compared to a linear Q-function approximator~\cite{sutton1998introduction}, a LMUT defines an ensemble of linear Q-function models, one for each partition cell. Since each Q-value prediction $Q^{UT}_{N}$ comes from a single linear model, the prediction can be explained by the feature weights of the model.

\begin{table}[htb]
    \begin{minipage}[t]{.76\textwidth}
    \centering
    \subfloat{
    \hspace*{-1cm} 
    \includegraphics[scale=.30]{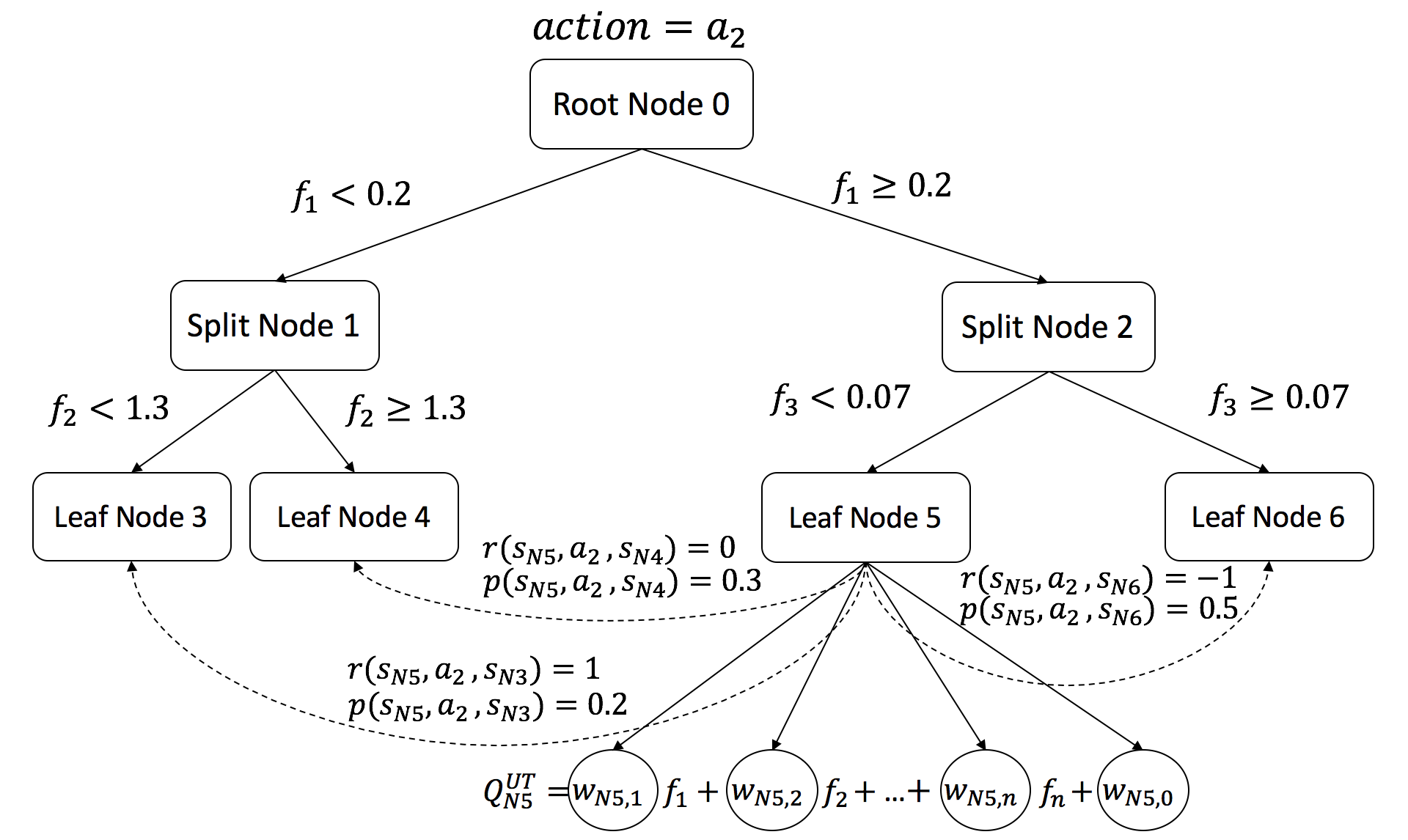}}
    \centering
    \captionsetup{justification=centering}
    \captionof{figure}{An example of Linear Model U-Tree (LMUT).}
    \label{fig:u-tree-structure}
     \end{minipage}%
    \begin{minipage}[t]{.25\textwidth}
    \centering
    \captionsetup{justification=centering}
    \caption{Partition Cell}
    \label{table:partition-cell}
    \begin{tabular}{|c|c|}
    \hline
    \multirow{2}{*}{\begin{tabular}[c]{@{}c@{}}Node \\ Name\end{tabular}} & \multirow{2}{*}{\begin{tabular}[c]{@{}c@{}}Partition\\ Cell\end{tabular}} \\
     &  \\ \hline
    \multirow{2}{*}{\begin{tabular}[c]{@{}c@{}}Leaf\\ Node 3\end{tabular}} & \multirow{2}{*}{\begin{tabular}[c]{@{}c@{}}$f_{1}<0.2$,\\ $f_{2}<1.3$\end{tabular}} \\
     &  \\ \hline
    \multirow{2}{*}{\begin{tabular}[c]{@{}c@{}}Leaf\\ Node 4\end{tabular}} & \multirow{2}{*}{\begin{tabular}[c]{@{}c@{}}$f_{1}<0.2$,\\ $f_{2}\geq1.3$\end{tabular}} \\
     &  \\ \hline
    \multirow{2}{*}{\begin{tabular}[c]{@{}c@{}}Leaf\\ Node 5\end{tabular}} & \multirow{2}{*}{\begin{tabular}[c]{@{}c@{}}$f_{1}\geq0.2$,\\ $f_{2}<0.07$\end{tabular}} \\
     &  \\ \hline
    \multirow{2}{*}{\begin{tabular}[c]{@{}c@{}}Leaf\\ Node 6\end{tabular}} & \multirow{2}{*}{\begin{tabular}[c]{@{}c@{}}$f_{1}\geq0.2$,\\ $f_{2}\geq0.07$\end{tabular}} \\
     &  \\ \hline
    \end{tabular}
    \end{minipage}
\end{table}


We now discuss how to train an LMUT. Similar to~\cite{uther1998tree}, we separate the training into two phases: 1) Data Gathering Phase and 2) Node Splitting Phase.

\subsection{Data Gathering Phase} 
Data Gathering Phase collects transitions on leaf nodes and prepares them for fitting linear models and splitting nodes.
Given an input transition $\transition$,  
we pass it through feature splits down to a leaf node.
As an option, an LMUT can dynamically build an MDP, in which case it updates transition probabilities, rewards and average Q values on the leaf nodes. 
The complete Data Gathering Phase process
is detailed in part I (the first for loop) of Algorithm~\ref{alg:LMUT-learning}.

\subsection{Node Splitting Phase}\label{section:node-splitting}
After node updating, LMUT scans the leaf nodes and updates their linear model with
Stochastic Gradient Descent (SGD).
If SGD achieves insufficient improvement on node N, LMUT determines a new split and adds the resulting leaves to the current partition cell.
For computational efficiency, our node splitting phase considers only a single split for each leaf given a single minibatch of new transitions. Part II   of Alg.\ref{alg:LMUT-learning} shows the detail of the node splitting phase. 
LMUT 
applies a minibatch stagewise fitting approach to learn linear models in the leaves of the U-tree. Like other stagewise approaches~\cite{landwehr2005logistic}, this approach provides smoothed weight estimates where nearby leaves tend to have similar weights.
We use Stochastic Gradient Descent to implement the weight updates. 

\begin{algorithm}[htbp]
\DontPrintSemicolon
\SetNoFillComment
\SetAlgoLined
    \KwIn{Transitions $\transition_{1},\ldots,\transition_{B}$; A LMUT with leaf nodes $N_1,\ldots,N_{L}$, each with weight vector $\bf{w_1},\ldots,\bf{w_{L}}$} 
    \kwhyper{$MinImprovement$, $minSplit$, $FlagMDP$}
    \;
    \tcc{Part I: Data Gathering Phase}
    \For{$t=1$ to $B$}
        {Find the partition cell on leaf node $N$ by $\signal_{t}, \action_{t}$ in $\transition_{t}$\;
        Add $\transition_{t}=\langle\signal_{t}, \action_{t}, \reward_{t}, {\signal_{t+1}}, \Qhat{t}{\signal_{t},\action_{t}} \rangle$ to transition set on $N$\;
        \If{$FlagMDP$ \tcc*{Update the Markov Decision Process}}
        {
            Map observation $(\signal_{t},\signal_{t+1})$ to state $(\state_{t},\state_{t+1})$ within partition cell of N\;
            Update Transitions Probability $P(\state_{t}, \action_{t},\state_{t+1}) = \frac{count(\state_{t}, \action_{t},\state_{t+1})+1}{\sum_{i}count(\state_{t}, \action_{t},\state_{i})+1}$\;
            
            Update Reward $R(\state_{t}, \action_{t}, \state_{t+1}) = \frac{R(\state_{t}, \action_{t}, \state_{t+1})*count(\state_{t}, \action_{t},\state_{t+1})+\reward_{t}}{count(\state_{t}, \action_{t},\state_{t+1})+1}$\;
            
            Compute $\Qtree{avg}{\state_{t}, \action_{t}}=\frac{\Qhat{t}{\state_{t}, \action_{t}}*count(\state_{t}, \action_{t})+\Qhat{t}{\signal_{t},\action_{t}}}{count(\state_{t}, \action_{t})+1}$\;
            Increment $count(\state_{t},\action_{t})$ and $count(\state_{t}, \action_{t},\state_{t+1})$ by 1\;
            }
        }
    \;
    \tcc{{Part II\bf}: Node Splitting Phase}
    \For{$i=1$ to $L$}{
    $w_i, err_i := \mbox{WeightUpdate}(\bf{\transition_{N_i}},\bf{w_i})$ \tcc*{Update the weights by SGD}
    \If{$err\leq MinImprovement$}{
        \For {distinction D in GetDistinction($N_{i}$)}{
            Split Node $N_{i}$ to FringeNodes by distinction D \;
            Compute distribution of Q function $\sigma_{N_{i}}(Q)$ on Node $N_{i}$\;
            \For {each FringeNodes $F$}{
                Compute distribution $\sigma_{F}(Q)$ on fringe node $F$\;
                \tcc{This function is discussed in Splitting Criterion}
                $p$ = SplittingCriterion($\sigma_{N_{i}}(Q)$, $\sigma_{F}(Q)$)\;
                \If{$p \geq minSplit$}{
                    $BestD = D$\;
                    $minSplit = p$\;
                }
            Remove all the fringe nodes\;
            }
        }
         \If{$BestD$}{
            Split Node $N_i$ by $BestD$ to define ChildNodes $N_{i,1},\ldots,N_{i,C}$\;
            Assign Transitions set $\bf\transition_{N_i}$ to ChildNodes\;
            \For{$c=1$ to $C$}{
            \tcc{Child Node weights are inherited from parent Node} 
            $w_{i,c} := w_i$\;
            $w_{i,c} , err_{i,c} := \mbox{WeightUpdate}(\bf\transition_{N_i,c},w_{i,c})$\; }
        }
    }
    
}
 \caption{Linear Model U-Tree Learning
 }\label{alg:LMUT-learning}
\end{algorithm}

\subsubsection{Stochastic Gradient Descent (SGD) Weight Updates}
is a straightforward well-established online weight learning method for a single linear regression model.
The weights and bias of linear regression on leaf node $N$ are updated by applying SGD over all Transitions assigned to $N$. 
For a transition $\transition_{t}= \langle\signal_{t}, \action_{t}, \reward_{t}, {\signal}_{t+1}, \hat{Q}(I_{t},a_{t}) \rangle$, we take $\signal_{t}$ as input and $\hat{Q}_{t} \equiv \hat{Q}(I_{t},\action_{t})$ as label. We build a separate LMUT for each action, so the linear model on $N$ is function of the $J$ state features: $\Qtree{}{I_{t}|w_{N},\action_{t}} = \sum_{j=1}^{J}I_{tj}w_{Nj}+w_{N0}$. We update the weights $w_N$ on leaf node $N$ by applying SGD 
with loss function 
$\mathcal{L}(w_{N}) = \sum_{t}1/2(\hat{Q}_{t}-\Qtree{}{I_{t}|w_{N},\action_{t}})^{2}$. The updates are computed with a single pass over each minibatch.

\begin{algorithm}[H]
\SetAlgoLined
\KwIn{Transitions $\transition_{1},\ldots,\transition_{m}$,
 node N = leaf node with weight vector $w_0$}
 \KwOut{updated weight vector $w$, training error $err$}
 \kwhyper{number of iterations $E$; step size $\alpha$}
 $w := w_0$ \;
 \For{
    $e$ = 1 to $E$
    }
    {\For{t=1 to $m$}
    {$w := w + \alpha \nabla_{w}\mathcal{L}(w)$ \;
    }
    }
Compute training error $err = 1/m \sum^{m}_{t=1}(\hat{Q}_{t}-\Qtree{}{I_{t}|w,\action_{t}})^{2}$
\caption{SGD Weight Update at a leaf nodes.}\label{alg:weight-update}
\end{algorithm}


\subsubsection{Splitting Criterion} is used to find the best split on the leaf node, if SGD achieves limited improvement. We have tried three splitting criteria including working response of SGD, Kolmogorov–Smirnov (KS) test and Variance Test. 
The first method 
aims to find the best split to improve working response of the parent linear model on the data for its children. But as reported in~\cite{landwehr2005logistic}, the final result becomes less intelligible.
The second method Kolmogorov–Smirnov (KS) test is a non-parametric statistical test that measures the differences in empirical cumulative distribution functions between the child data. 
The final Variance criterion selects a split that generates child nodes whose Q values contain the least variance. 
The idea is similar to the variance reduction method applied in CART tree.
Like ~\cite{uther1998tree}, we found that Variance test works well with less time complexity than KS test ($O(n)$ v.s. $O(n^{2})$),  so we select Variance test as the splitting criterion. Exploring the different possible splits efficiently is the main scalability challenge in LMUT learning (cf.~\cite{uther1998tree}).


\section{Empirical Evaluation}

    We evaluate the mimic performance of LMUT by comparing it with five other baseline methods under three evaluation environments.
    Empirical evaluation measures both regression and game playing matches under experience training and active play learning. 
    
    \subsection{Evaluation Environment}
    The evaluation environments include \textbf{Mountain Car}, \textbf{Cart Pole} and \textbf{Flappy Bird}. Our environments are simulated by OpenAI Gym toolkit~\cite{brockman2016openai}.
    Mountain Car and Cart Pole are two benchmark tasks for reinforcement learning~\cite{riedmiller2005neural}. Mountain Car is about accelerating a car to the top of the hill and Cart Pole is about balancing a pole in the upright position.
    Mountain Car and Cart Pole have a discrete action space and 
    continuous feature space.
    Flappy Bird is a mobile game that controls a bird to fly between pipes.
    Flappy Bird has two discrete actions, and its observation consists of four consecutive images~\cite{Mnih2015}. We follow the Deep Q-Learning (DQN) method to play this game. 
    During the image preprocessing, the input images are first rescaled to 80*80, transferred to gray image and then binary images. With 6,400 features, the state space of Flappy Bird is substantially more complex than that for Cart Pole and Mountain Car. 
        
    \subsection{Baseline Methods}
    
    \textbf{CART} is our first baseline method, where we fit the input/output training pairs $(\langle\signal,\action\rangle,\Qhat{}{\signal,\action})$ into a CART regression tree~\cite{loh2011classification}. 
    A CART tree predicts the mean of sample $Q$ on each leaf node.
    M5~\cite{Quinlan1992} is a tree training algorithm with more generalization ability. It first constructs a piecewise constant tree 
    and then prunes to build a linear regression model for the instances in each leaf node.  
    The WEKA toolkit~\cite{hall2009weka} provides an implementation of M5. We include M5 with Regression-Tree option (\textbf{M5-RT}) and M5 tree with Model-Tree option (\textbf{M5-MT}) in our baselines. M5-MT builds a linear function on each leaf node, while M5-RT has only a constant value.
    To compare the online training performance, a recently proposed online learning model \textbf{Fast Incremental Model Tree (FIMT)}~\cite{ikonomovska2011learning} is applied. Similar to M5-MT, it builds a linear model tree, but can perform explicit change detection and informed adaption for evolving data stream. We experiment with a basic version of FIMT and an advanced version with {\bf Adaptive Filters} on leaf nodes (named \textbf{FIMT-AF}).
    
    
    \subsection{Fidelity: Regression Performance}
    We evaluate how well our LMUT approximates the soft output ($\hat{Q}$ values) from Q function in a Deep Q-Network (DQN).
    We report the standard regression metrics {\bf Mean Absolute Error (MAE)}, and {\bf Root Mean Square Error (RMSE)}.
    Under the {\em Experience Training} setting, we compare the performance of CART, M5-RT, M5-MT, FIMT and FIMT-AF with our LMUT. The dataset sizes are 150K transitions for Mountain Car, 70K transitions for Car Pole, and 20K transitions for Flappy Bird. Because of the high dimensionality of the Flappy Bird state space, 32GB main memory fits only 20K transitions. 
    Given an experience training dataset, we apply 10 fold cross evaluation to train and test our model.
    For the {\em Active Play} setting, batch training algorithms like CART and M5 are not applicable, so we experiment only with online methods, including FIMT, FIMT-AF and LMUT. We first train the mimic models with 30k consecutive transitions from evaluation environments,
    and evaluate them with
    another 10k transitions. 
    The result for the three evaluation environments are shown in Table~\ref{table:mc-result}, Table~\ref{table:cp-result} and Table~\ref{table:fb-result}.
    A t-test demonstrates that the differences between the results of LMUT and the results of other models are significant ($p <5\%$).

    Compared to the other two online learning methods (FIMT and FIMT-AF), LMUT achieves a better fit to the neural net predictions with a much smaller model tree,
    especially in the active play online setting.
    This is because both FIMT and FIMT-AF update their model tree continuously after each datum,
    whereas LMUT fits minibatches of data at each leaf.
    Neither FIMT nor FIMT-AF terminate on high-dimensional data.\footnote{For example, in  the Flappy Bird environment,  FIMT takes 29 minutes and 10.8GB main memory to process 10 transitions on a machine using i7-6700HQ CPU.} 
    So we {\it omit} the result of applying FIMT and FIMT-AF in the Flappy Bird environment.
    We observe that the CART tree model has significantly more leaves than our LMUT, but not better fit to the DQN than M5-RT, M5-MT and LMUT, which suggests overfitting.
    In the Mountain Car and Flappy Bird environments, model tree batch learning (M5-RT and M5-MT) performs better than LMUT, while LMUT achieves comparable fidelity, and leads in the Cart Pole environment. 
    In conclusion, (1) our LMUT learning algorithm outperforms the state-of-the-art online model tree learner FIMT. (2) Although LMUT is an online learning method, it showed competitive performance to batch methods even in the batch setting. 
    
    \begin{table}[htbp]
    \centering
    \begin{minipage}[t]{.5\textwidth}\centering
    \caption{Result of Mountain Car}
    \label{table:mc-result}
    \resizebox{0.9\columnwidth}{!}{
        \begin{tabular}{|c|c|ccc|}
        \hline
        \multicolumn{2}{|c|}{\multirow{2}{*}{Method}} & \multicolumn{3}{c|}{Evaluation Metrics} \\ \cline{3-5} 
        \multicolumn{2}{|c|}{} & MAE & RMSE & Leaves \\ \hline
        \multirow{6}{*}{\begin{tabular}[c]{@{}c@{}}Expe-\\ rience\\ Train-\\ing\\\end{tabular}} & CART & 0.284 & 0.548  & 1772.4  \\
         & M5-RT & 0.265 & 0.366 & 779.5  \\
         & {\bf M5-MT} & {\bf 0.183}  & {\bf 0.236}  & 240.3  \\
         & FIMT & 3.766 & 5.182 & 4012.2  \\
         & FIMT-AF & 2.760 & 3.978& 3916.9  \\
         & LMUT & 0.467 & 0.944 & 620.7  \\ \hline
        \multirow{3}{*}{\begin{tabular}[c]{@{}c@{}}Active\\ Play\end{tabular}} & FIMT & 3.735 & 5.002 & 1020.8 \\
         & FIMT-AF  & 2.312 & 3.704 & 712.4  \\
         & LMUT & 0.475 & 1.015 & 453.0  \\ \hline
        \end{tabular}
    }
    \end{minipage}\hfill
    \begin{minipage}[t]{.5\textwidth}
    \centering
    \caption{Result of Cart Pole}
    \label{table:cp-result}
    \resizebox{0.93\columnwidth}{!}{
        \begin{tabular}{|c|c|ccc|}
        \hline
        \multicolumn{2}{|c|}{\multirow{2}{*}{Method}} & \multicolumn{3}{c|}{Evaluation Metrics} \\ \cline{3-5} 
        \multicolumn{2}{|c|}{} & MAE & RMSE & Leaves \\ \hline
        \multirow{6}{*}{\begin{tabular}[c]{@{}c@{}}Expe-\\ rience\\ Train-\\ing\\\end{tabular}} &  CART & 15.973 & 34.441  & 55531.4  \\
         & M5-RT & 25.744 & 48.763  & 614.9  \\
         & M5-MT & 19.062 & 37.231  & 155.1  \\
         & FIMT & 43.454 & 65.990  & 6626.1 \\
         & FIMT-AF  & 31.777 & 50.645& 4537.6  \\
         & {\bf LMUT} & {\bf13.825} & {\bf27.404}  & 658.2  \\ \hline
        \multirow{3}{*}{\begin{tabular}[c]{@{}c@{}}Active\\ Play\end{tabular}} & FIMT &32.744 & 62.862 & 2195.0  \\
         & FIMT-AF  & 28.981 & 51.592  & 1488.9  \\
         & LMUT & 14.230 & 43.841 & 416.2  \\ \hline
        \end{tabular}
    }
    \end{minipage}
    \end{table}
    
    \begin{table}[htbp]
    \centering
    \begin{minipage}[t]{.5\textwidth}
    \centering
    \caption{Result of Flappy Bird}
    \label{table:fb-result}
    \resizebox{0.88\columnwidth}{!}{
        \begin{tabular}{|c|c|ccc|}
        \hline
        \multicolumn{2}{|c|}{\multirow{2}{*}{Method}} & \multicolumn{3}{c|}{Evaluation Metrics} \\ \cline{3-5} 
        \multicolumn{2}{|c|}{} & MAE & RMSE & Leaves \\ \hline
        \multirow{4}{*}{\begin{tabular}[c]{@{}c@{}}Expe-\\ rience\\ Train-\\ing\\\end{tabular}} &
        CART & 0.018 & 0.036  & 700.3 \\
        &M5-RT & 0.027 & 0.041  & 226.1 \\
        &{\bf M5-MT} & {\bf 0.016} & {\bf 0.030} & 412.6  \\
        &LMUT & 0.019 & 0.043  & 578.5 \\ \hline
        \multirow{2}{*}{\begin{tabular}[c]{@{}c@{}}Active\\ Play \end{tabular}}
        &\multirow{2}{*}{LMUT} & \multirow{2}{*}{0.024} & \multirow{2}{*}{0.050} &  \multirow{2}{*}{229.0} \\
        &  &  &  & \\\hline
        \end{tabular}
    }
    \end{minipage}%
    \begin{minipage}[t]{.5\textwidth}
    \centering
    \subfloat{
    \includegraphics[scale=.2]{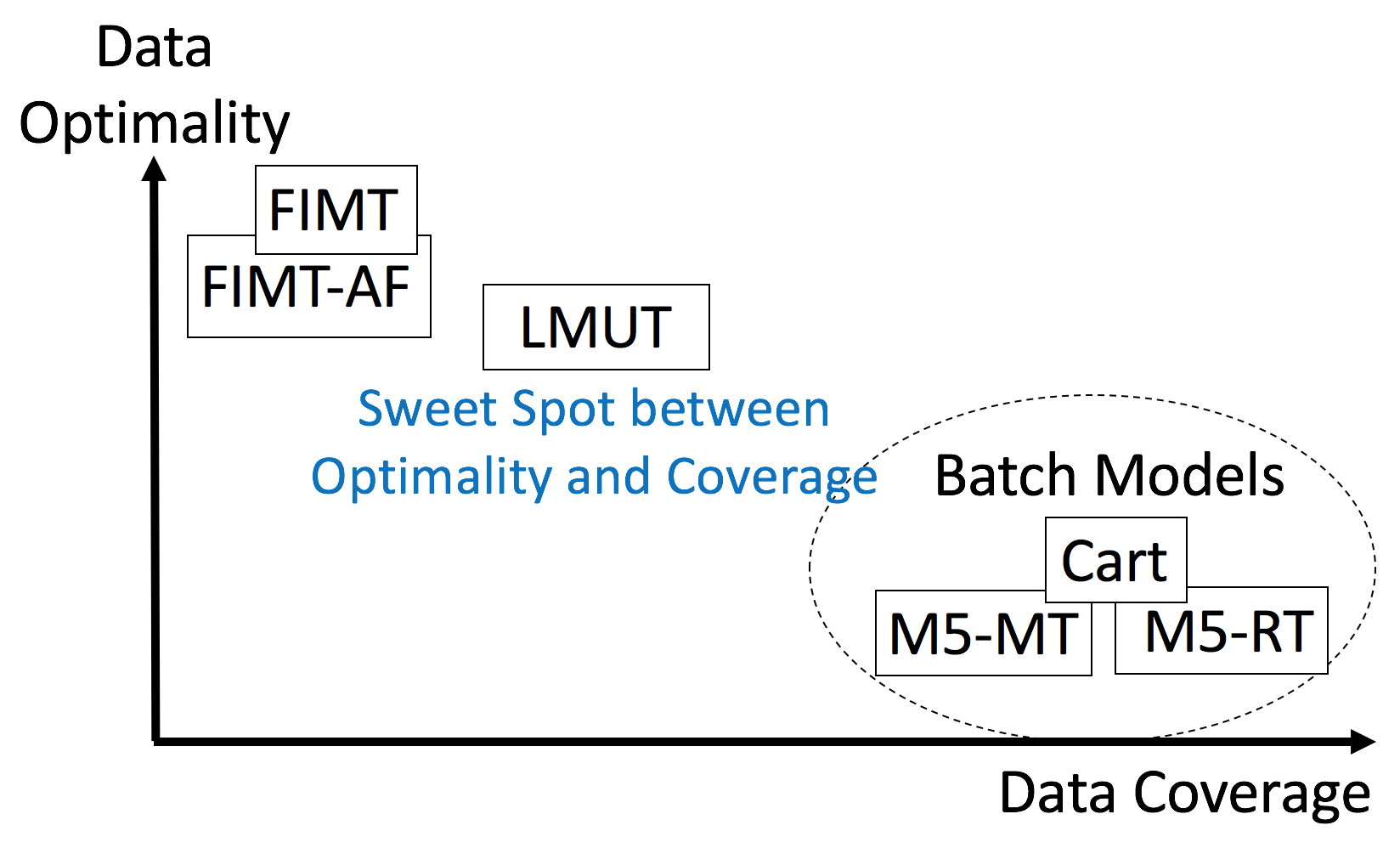}}
    \captionof{figure}{Coverage v.s. Optimality
    }
    \label{fig:balanace-data}
    \end{minipage}%
    \end{table}
    
\begin{figure}[htbp]
    \begin{minipage}{.33\textwidth}
    \centering
    \subfloat{\includegraphics[scale=.185]{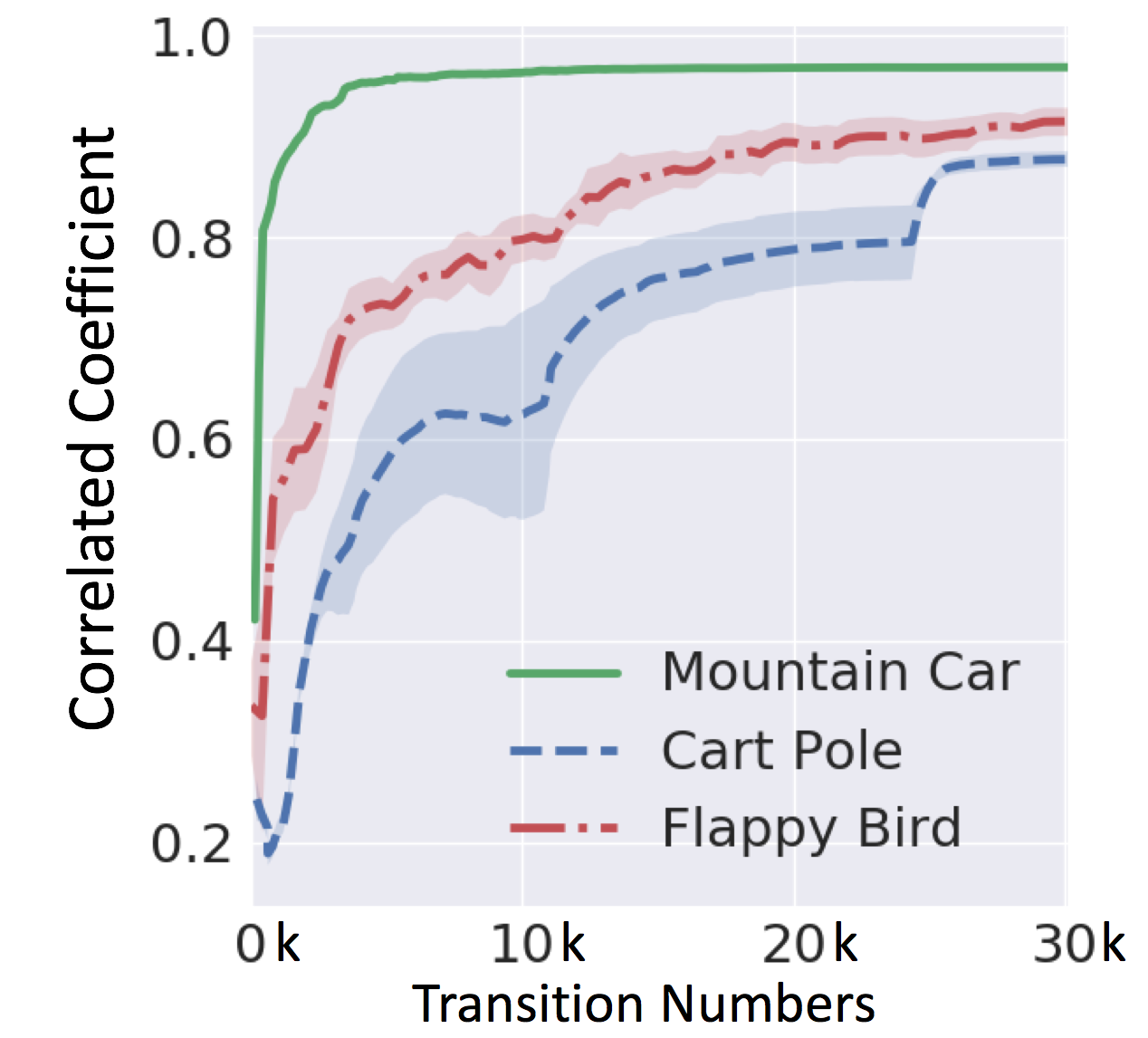}}
    \end{minipage}%
    \begin{minipage}{.33\textwidth}
    \centering
    \subfloat{\includegraphics[scale=.185]{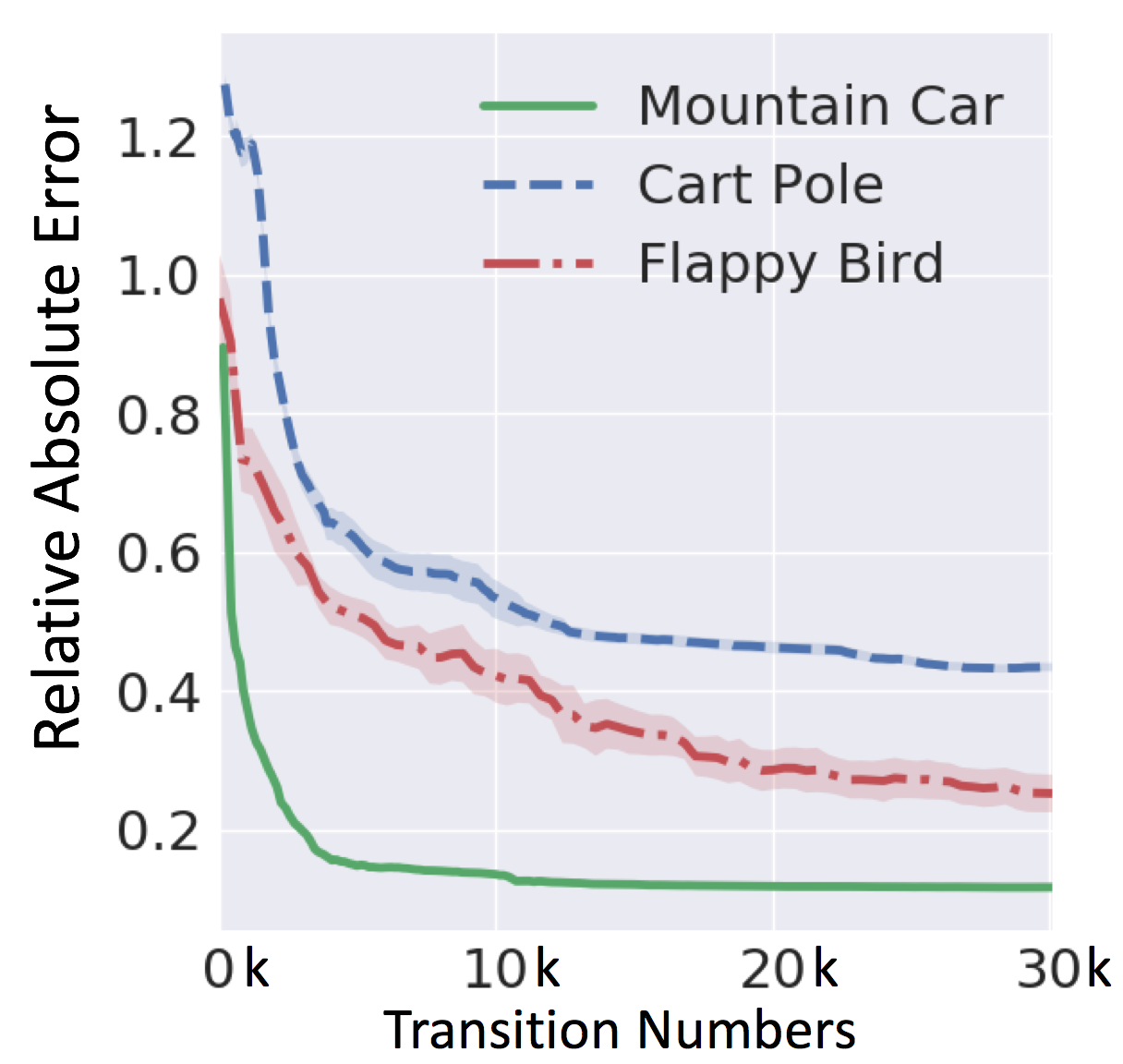}}
    \end{minipage}%
    \begin{minipage}{.33\textwidth}
    \centering
    \subfloat{\includegraphics[scale=.185]{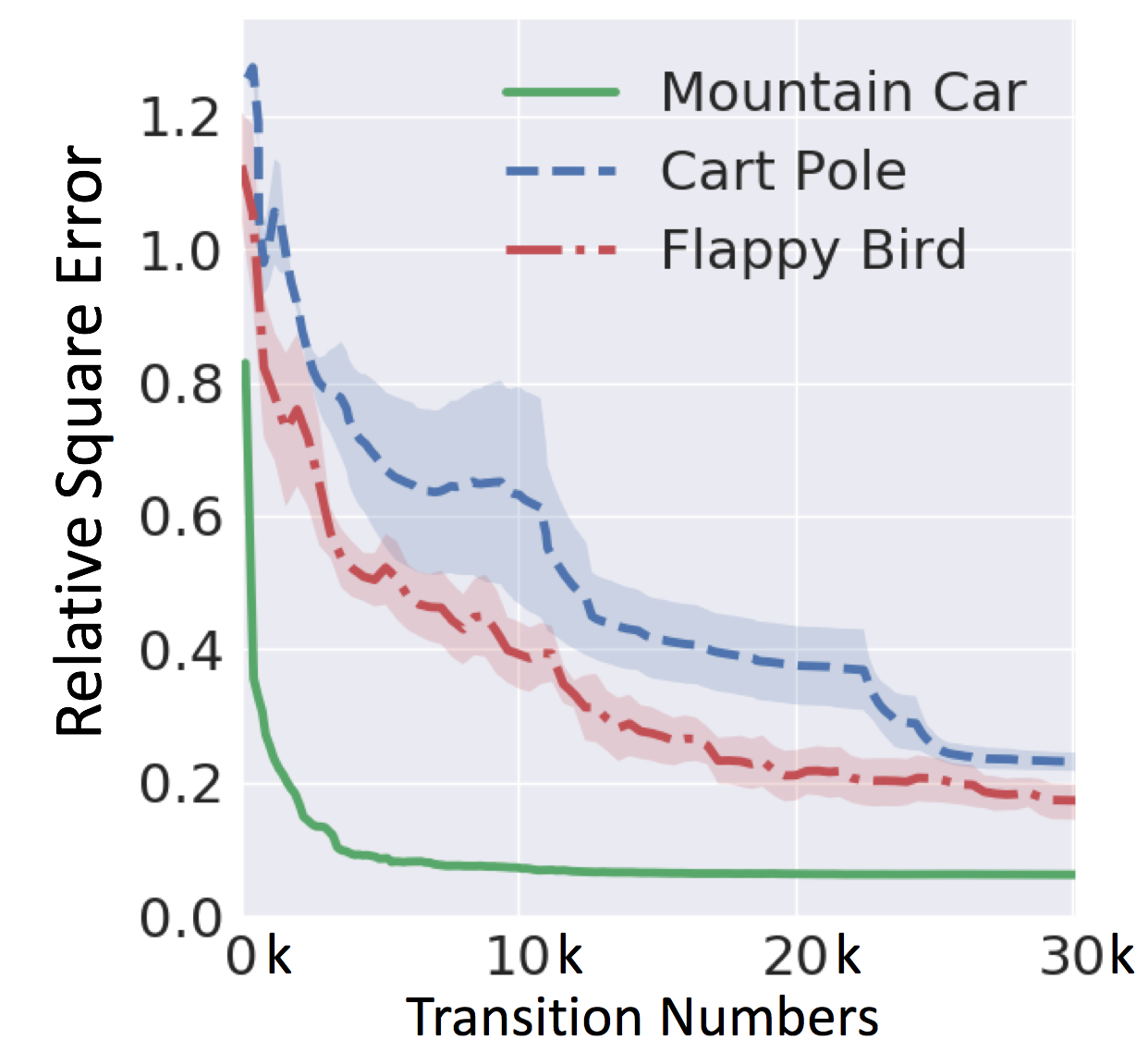}}
    \end{minipage}
    \caption{Consecutive Testing of LMUT}
    \label{fig:lmut-result}
    \end{figure}

    {\em Learning Curves.} We apply consecutive testing~\cite{ikonomovska2011learning} to analyze the performance of LMUT learning in more detail. 
    We compute the correlation and testing error of LMUT 
    as more transitions for learning are provided (From 0 to 30k) 
    under the active play setting.
    To adjust the error scale across different game environments, we use {\bf Relative Absolute Error (RAE)} and {\bf Relative Square Error (RSE)}.
    We repeat the experiment 10 times and plot the shallow graph in Figure~\ref{fig:lmut-result}.
    In the Mountain Car environment, LMUT converges quickly with its performance increasing smoothly in 5k transitions. But for complex environments like Cart Pole and Flappy Bird, the evaluation metrics fluctuate during the learning process but will approximate to the optimum within 30k transitions. 

    \subsection{Matching Game Playing Performance}
    We now evaluate how well a model mimics Q functions in DQN by directly playing the games with them and computing the average reward per episode. (The games in OpenAI Gym toolkit are divided into episodes that start when a game begins and terminate when: (1) the player reaches the goal, (2) fails for a fixed number of times or (3) the game time passes a preset threshold). 
    Specifically, given an input signal $\signal_{t}$, we obtain $Q$ values from mimic models and select an action $\action_{t}=\max_{\action}Q(\signal_{t},\action)$. By executing $\action_{t}$ in the current game environment, we receive a reward $\reward_{t}$ and next observation signal $\signal_{t+1}$. This process is repeated until a game episode terminates. 
    This experiment uses {\bf Average Reward Per Episodes (ARPE)}, a common evaluation metric that has been applied by both DRL models~\cite{Mnih2015} and OpenAI Gym tookit~\cite{brockman2016openai}, to evaluate mimic models.
    In the {\em Experience Training} setting, the play performance of CART, M5-RT, M5-MT, FIMT, FIMT-AF and our LMUT are evaluated and compared by partial 10-fold cross evaluation, where 
    we select 9 sections of data to train the mimic models and test them by directly playing another 100 games. 
    For the {\em Active play}, only the online methods FIMT and FIMT-AF are compared, without the Flappy Bird environment (as discussed in Section 5.3). Here we train the mimic models with 30k transitions, and test them in another 100 games.

    The result of game playing performance is shown in Table~\ref{table:average-reward}.
    We first experiment with learning a Continuous U-Tree (CUT) 
    {\em directly using reinforcement learning} \cite{uther1998tree} instead of mimic learning. 
    CUT converges slowly with limited performance, especially in the high-dimensional  Flappy Bird environment. This shows the difficulty of directly constructing a tree model from the environment. 
    
    We find that {\em among all mimic methods, LMUT achieves the Game Play Performance APER closest to the DQN.  
    Although the batch learning models have strong fidelity in regression,
    they do not perform as well in game playing as the DQN.}
    Game playing observation shows that 
    the batch learning models (CART, M5-RT, M5-MT) are
    likely to choose sub-optimal actions under some key scenarios (e.g.,
    when a pole tilts to one side with high velocity in Cart Pole.). This is because the neural net controller selects many sub-optimal actions at the beginning of training, so the early training experience contains many sub-optimal state-action pairs.  
    The batch models fit the entire training experience equally, 
    while our LMUT 
    fits more closely the most recently generated transitions from a mature controller. More recent transitions tend to correspond to optimal actions.
    The FIMT algorithms keep adapting to the most recent input only, and fail to build adequate linear models on their leaf nodes. Compared to them,  LMUT achieves a sweet spot between optimality and coverage (Figure~\ref{fig:balanace-data}).
    \begin{table}[htb!]
    \centering
    \caption{Game Playing Performance}
    \label{table:average-reward}
    \begin{tabular}{|c|c|c|c|c|}
    \hline
    \multicolumn{2}{|c|}{\multirow{2}{*}{Model}} & \multicolumn{3}{c|}{Game Environment} \\ \cline{3-5} 
    \multicolumn{2}{|c|}{} & Mountain Car & Cart Pole & Flappy Bird \\ \hline
    {\it Deep Model} & \textit{DQN} & {\it-126.43} & {\it175.52} & {\it123.42 } \\ \hline
    Basic Model & CUT & -200.00 & 20.93 & 78.51  \\ \hline
    \multirow{6}{*}{\begin{tabular}[c]{@{}c@{}}Experience\\ Training\end{tabular}} & CART & -157.19 & 100.52  & 79.13 \\
     & M5-RT & -200.00 & 65.59 & 42.14 \\
     & M5-MT & -178.72 & 49.99& 78.26 \\
     & FIMT & -190.41 & 42.88 & N/A \\
     & FIMT-AF & -197.22 & 37.25 & N/A  \\
     & { LMUT} & {-154.57} & {145.80} & {97.62} \\ \hline
    \multirow{3}{*}{\begin{tabular}[c]{@{}c@{}}Active\\ Play\end{tabular}} 
     & FIMT & -189.29 & 40.54 & N/A  \\
     & FIMT-AF & -196.86 & 29.05 & N/A  \\
     & {\bf LMUT} & {\bf -149.91} & {\bf \bf147.91 }  & {\bf103.32} \\ \hline
    \end{tabular}
    \end{table}

\section{Interpretability}
In this section, we discuss how to interpret a DRL model through analyzing the knowledge stored in the transparent tree structure of LMUT: computing feature influence, analyzing the extracted rules and highlighting the super-pixels.


\subsection{Feature Influence}
Feature importance is one of the most common interpretation tools for tree-based models~\cite{che2016interpretable,wu2017beyond}. In a LMUT model, feature values are used as splitting thresholds to form partition cells for input signals. 
We evaluate the influence of a splitting feature by the 
total variance reduction of the Q values. 
The absolute weight values from linear regression provide extra knowledge of feature importance.
So we compute a weight importance rate and multiply it by Variance Reduction, and measure the influence of splitting feature $f$ on node $N$ by:
\begin{equation}
    \it{Inf}_{f}^{N} =(1+\frac{|w_{Nf}|^{2}}{\sum_{j=1}^{J}|w_{Nj}|^{2}})(var_{N} 
    -\sum_{c=1}^{C}\frac{Num_{c}}{\sum_{i=1}^{C}Num_{i}}var_{c}),
\end{equation}
where $w_{Nf}$ is the weight of feature $f$ on node N, $Num_{c}$ is the number of Q values on node $c$ and $var_{N}$ is the variance of Q values on node $N$.
We quantify the influence of a splitting feature $\it{Inf}_{f}$ by summing $\it{Inf}_{f}^{N}$ for all nodes $N$ split by $f$ in our LMUT.
For 
Mountain Car and Cart Pole, we report the feature influences in table~\ref{table:feature-influence}. 
The most important feature for Mountain Car and Cart Pole are Velocity and 
Pole Angle respectively, which matches the common understanding of the domains.
For Flappy Bird whose observations are 80*80 images, LMUT uses pixels as splitting features. Figure~\ref{fig:play-performance} illustrates the pixels with above-average feature influences $\it{Inf}_{f}>0.008$ (the mean of all feature influences).
The most influential pixels are located on the top left where the bird is likely to stay, which reflects the importance of locating the bird.
\begin{table}[htbp]
    \begin{minipage}[t]{.5\textwidth}
    \centering
    \caption{Feature Influence}
    \label{table:feature-influence}
    \begin{tabular}{|c|cc|}
    \hline
     & Feature & Influence \\ \hline
    \multicolumn{1}{|c|}{\multirow{2}{*}{\begin{tabular}[c]{@{}c@{}}Mountain \\ Car\end{tabular}}} & Velocity & 376.86 \\
    \multicolumn{1}{|c|}{} & Position & 171.28 \\ \hline
    \multirow{4}{*}{\begin{tabular}[c]{@{}c@{}}Cart \\ Pole\end{tabular}} & Pole Angle &  30541.54 \\
     & Cart Velocity & 8087.68 \\
     & Cart Position &   7171.71 \\
     & Pole Velocity At Tip &  2953.73 \\ \hline
    \end{tabular}
    \end{minipage}%
    \begin{minipage}[t]{.5\textwidth}
    \centering
    \subfloat{
    \includegraphics[scale=.26]{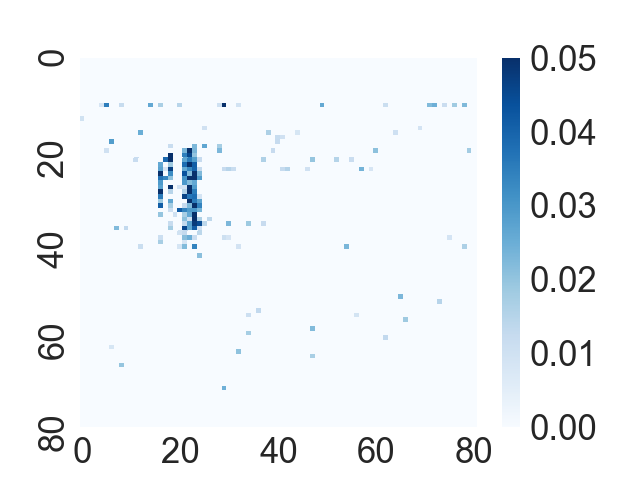}}
    \centering
    \captionsetup{justification=centering}
    \captionof{figure}{Super pixels in Flappy Bird \\
    }
    \label{fig:play-performance}
     \end{minipage}%
\end{table}

\subsection{Rule Extraction}

\begin{figure}[htbp]
    \begin{minipage}{.33\textwidth}
    \centering
    \subfloat{\label{fig:mc-a}\includegraphics[scale=.135]{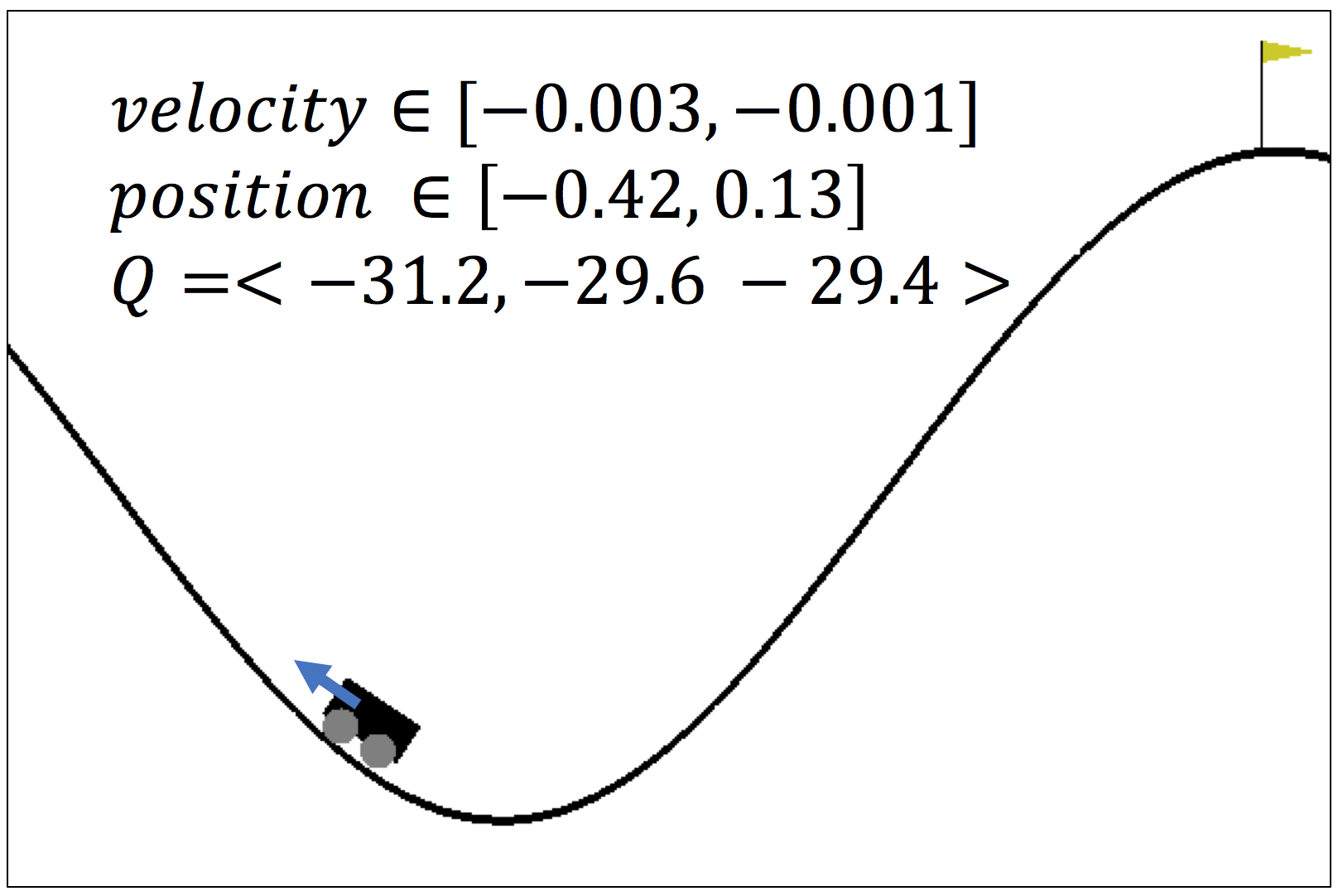}}
    \end{minipage}%
    \begin{minipage}{.33\textwidth}
    \centering
    \subfloat{\label{fig:mc-b}\includegraphics[scale=.135]{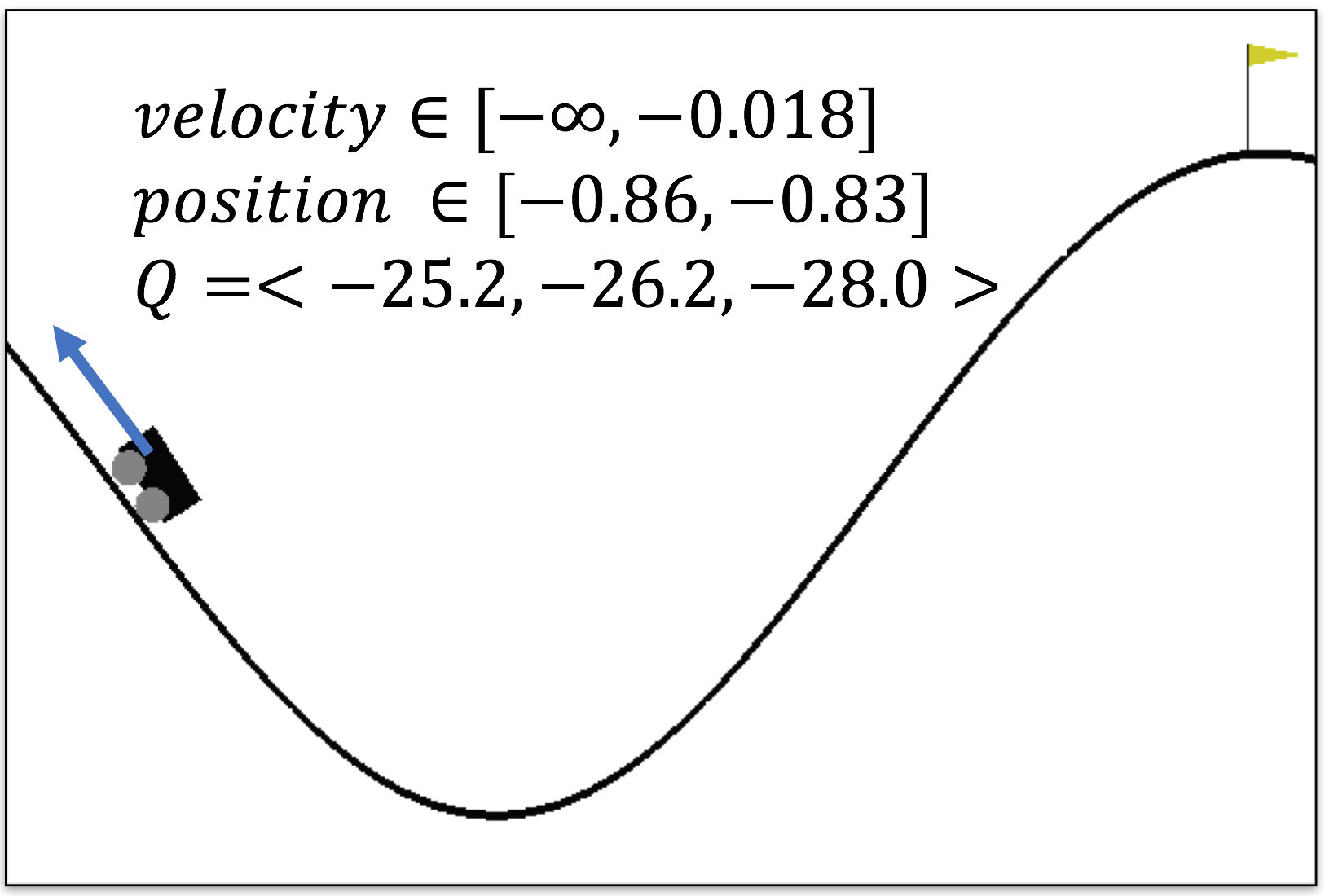}}
    \end{minipage}%
    \begin{minipage}{.33\textwidth}
    \centering
    \subfloat{\label{fig:mc-c}\includegraphics[scale=.135]{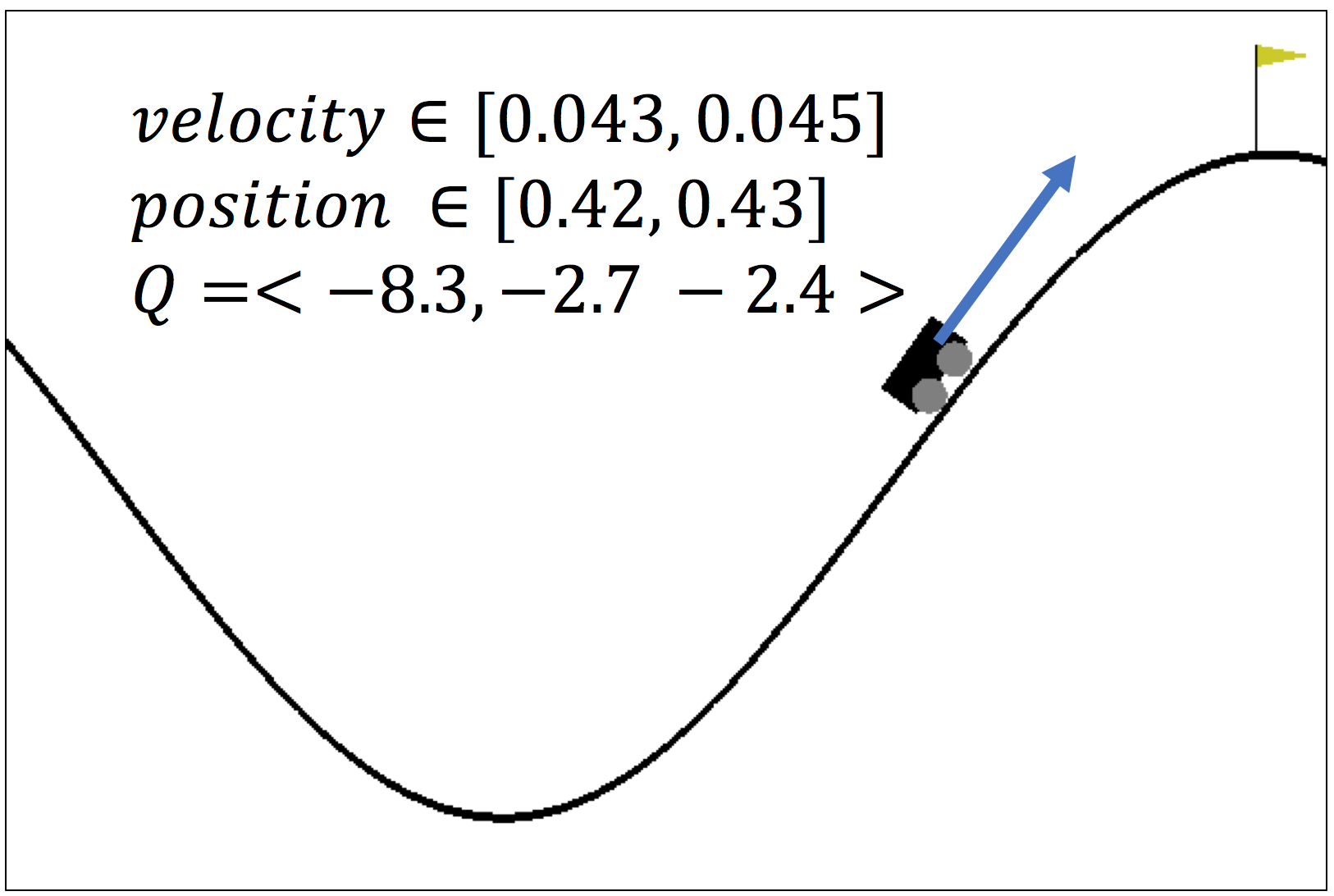}}
    \end{minipage}
    \par\smallskip
    \begin{minipage}{.33\textwidth}
    \centering
    \subfloat{\label{fig:cp-a}\includegraphics[scale=.135]{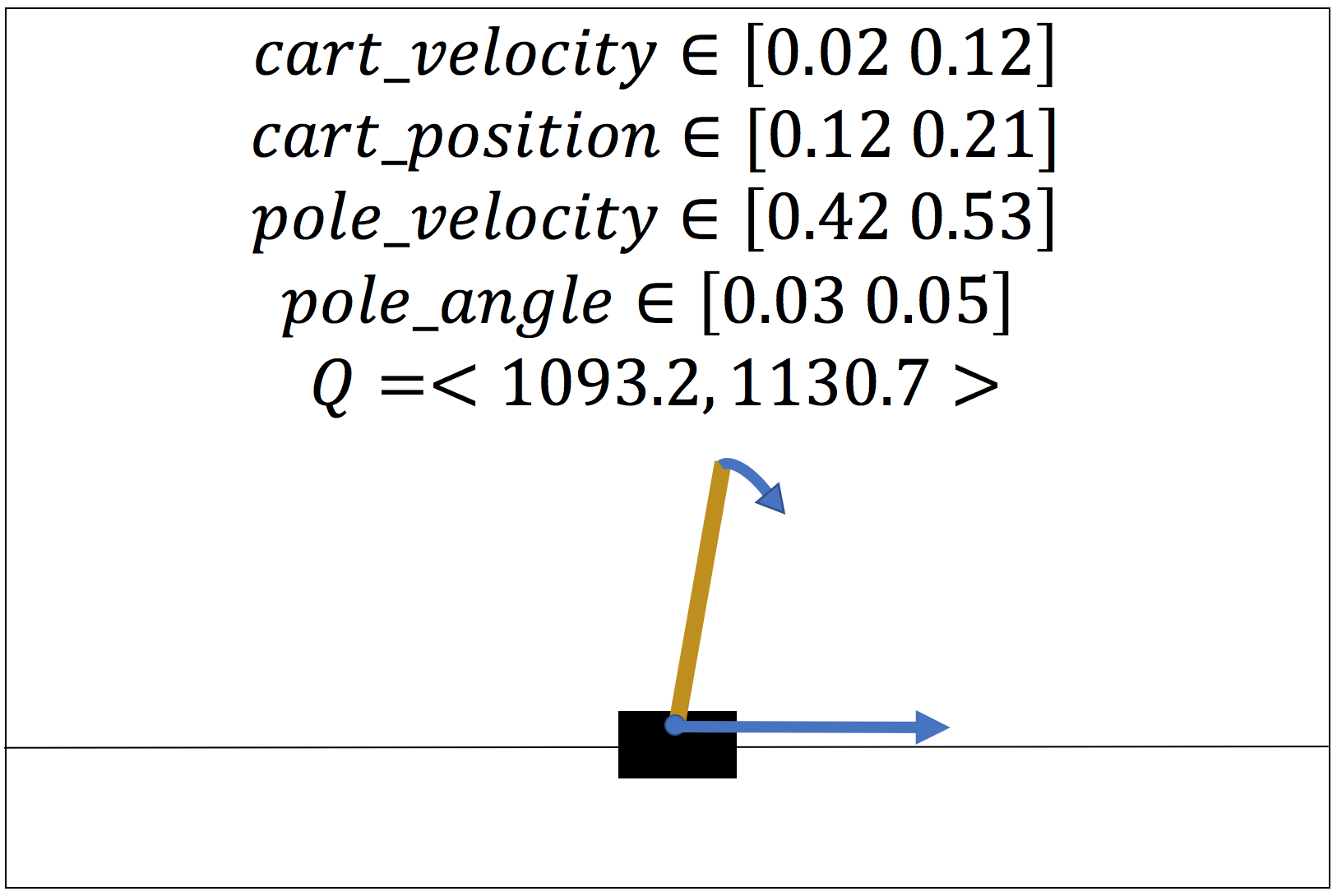}}
    \end{minipage}%
    \begin{minipage}{.33\textwidth}
    \centering
    \subfloat{\label{fig:cp-b}\includegraphics[scale=.135]{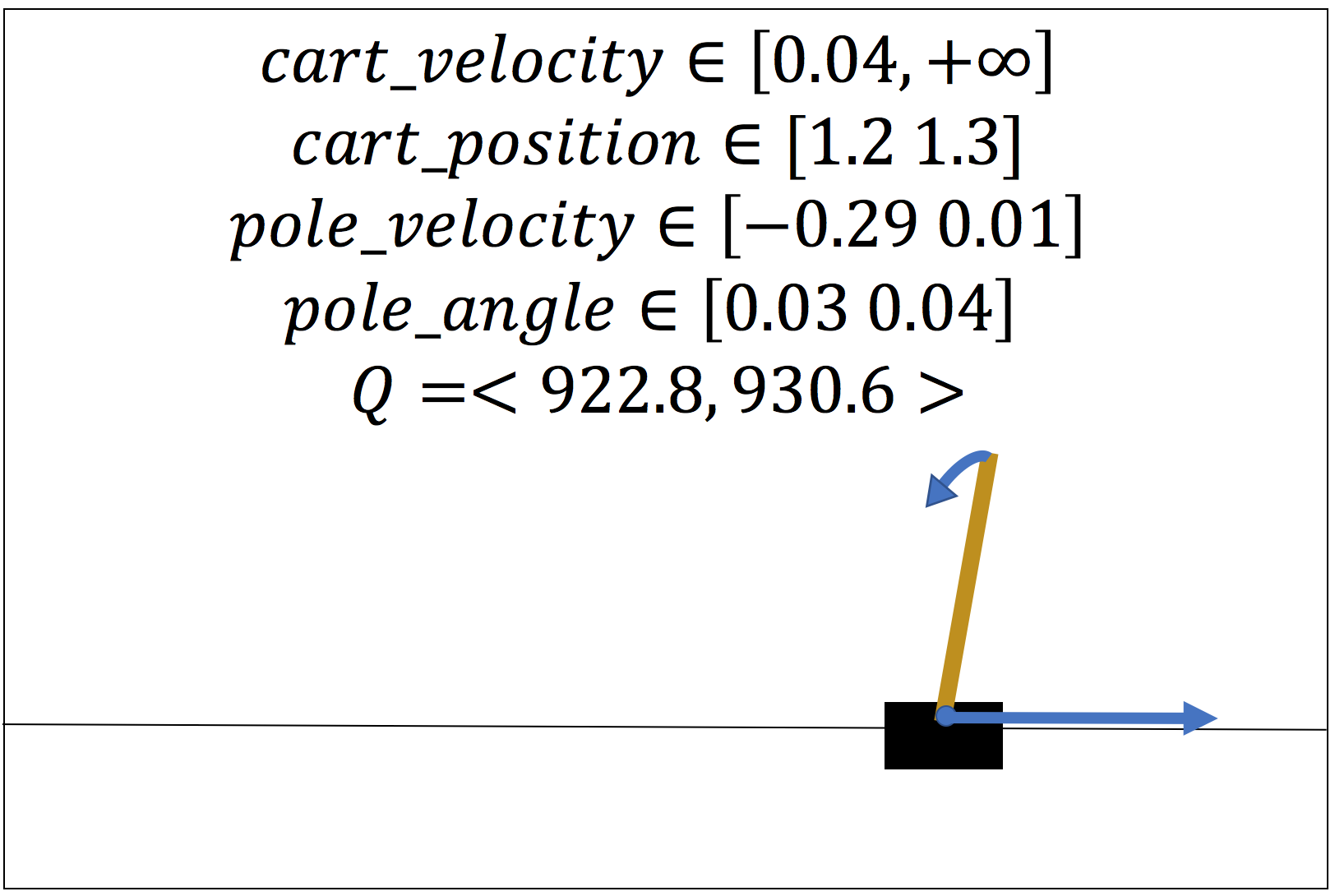}}
    \end{minipage}%
    \begin{minipage}{.33\textwidth}
    \centering
    \subfloat{\label{fig:cp-c}\includegraphics[scale=.135]{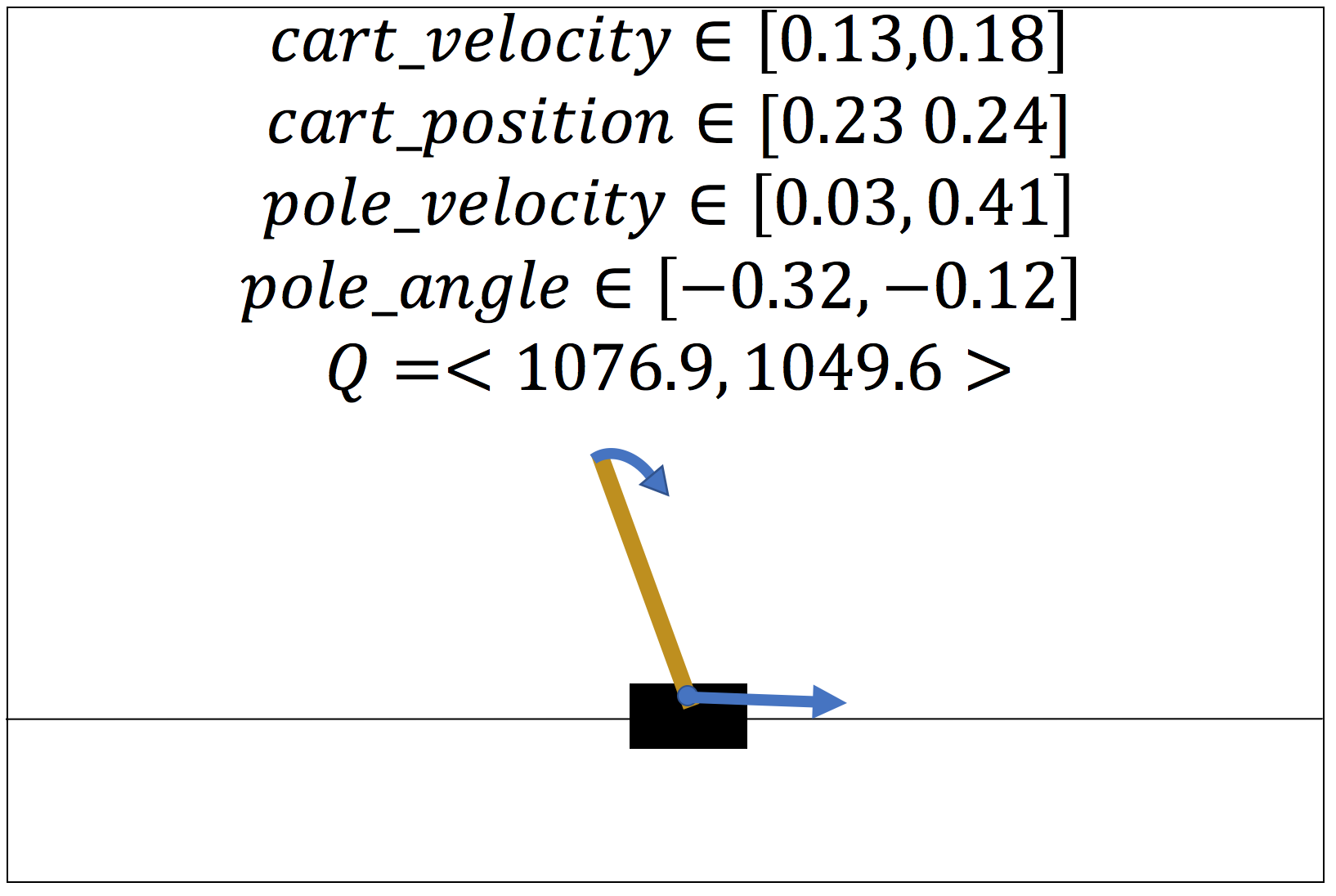}}
    \end{minipage}
    \caption{Examples of Rule Extraction for Mountain Car and Cart Pole. 
    }
    \label{fig:mc-rules}
\end{figure}

Rule extraction is a common method to extract knowledge from tree  models~\cite{dancey2007logistic,craven1996extracting,boz2002extracting}. We extract and analyze rules for the Mountain Car and Cart Pole environment.
Figure~\ref{fig:mc-rules} (top) shows three typical examples of extracted rules in Mountain Car environment. The rules are presented in the form of partition cells (constructed by the splitting features in LMUT). Each cell contains the range of velocity, position and a Q vector ($\mathbf{Q}=\langle Q_{move \_left}, Q_{no\_ push}, Q_{move\_ right}\rangle$) representing the average $Q$-value in the cell. 
The top left example is a state where the cart is moving toward the left hill with very small velocity. The extracted rule suggests pushing right ($Q_{move\_ right}$ has the largest value -29.4): the cart is almost stopped on the left, and by pushing right, 
it can increase its momentum (or Kinetic Energy).
The top middle example illustrates a state where the car is approaching the top of the left hill with larger left side velocity (compared to the first example). In this case, however, the cart should be pushed left ($Q_{move\_ left}=-25.2$ is the largest), in order to store more Gravitational Potential Energy and prepare for the final rush to the target. The rush will lead to the state shown in the top right image,  where the cart is rushing up the right hill. In this state, the cart should be pushed right to reach the target.
We also observe if the cart can reach the target in fewer steps, its Q values are larger. 

Figure~\ref{fig:mc-rules} (bottom) shows three examples of extracted rules in the Cart Pole environment, where each cell contains the scope of cart position, cart velocity, pole angle, pole velocity and a Q vector ($\mathbf{Q}=\langle Q_{push \_left}, Q_{push\_ right}\rangle$).
The key for Cart Pole is using inertia and acceleration to balance the pole.
In the bottom left example, the cart should be pushed right,
according to the rules ($Q_{push\_ right}>Q_{push \_left}$),  if the pole tilts to the right with a velocity less than 0.5. 
A similar scenario happens in the second example, where the pole is also tilting to the right but has velocity towards the left. We should push right ($Q_{push\_ right}>Q_{push \_left}$) to maintain this trend even if the cart is close to the right side border, which makes its Q values smaller than that in the first example. The third example describes 
a case where a pole tilts to the left with velocity towards the right. This time we need a left acceleration so the model selects pushing cart left ($Q_{push\_ right}<Q_{push \_left}$).

    \subsection{Super-pixel Explanation}
\begin{figure}[htb]
    \centering
        \begin{minipage}{.23\textwidth}
        \centering
        \subfloat{\label{fig:flappy-bird-split-group1:a}\includegraphics[scale=.18]{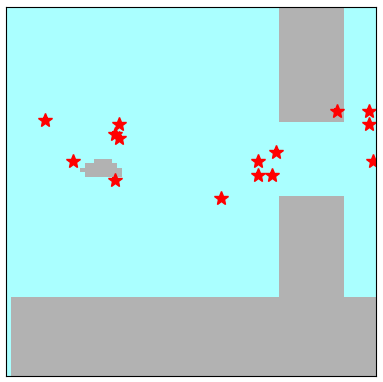}}
        \end{minipage}%
        \begin{minipage}{.23\textwidth}
        \centering
        \subfloat{\label{fig:flappy-bird-split-group1:b}\includegraphics[scale=.18]{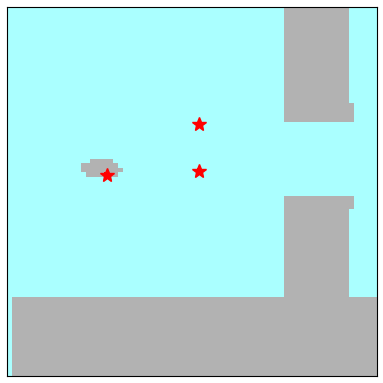}}
        \end{minipage}%
        \begin{minipage}{.23\textwidth}
        \centering
        \subfloat{\label{fig:flappy-bird-split-group1:c}\includegraphics[scale=.18]{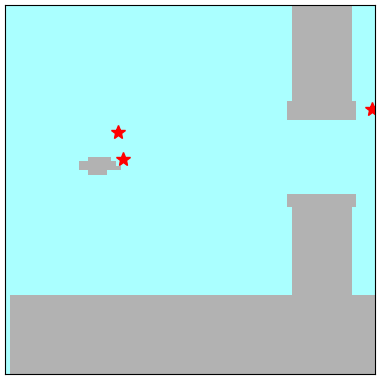}}
        \end{minipage}%
        \begin{minipage}{.23\textwidth}
        \centering
        \subfloat{\label{fig:flappy-bird-split-group1:d}\includegraphics[scale=.18]{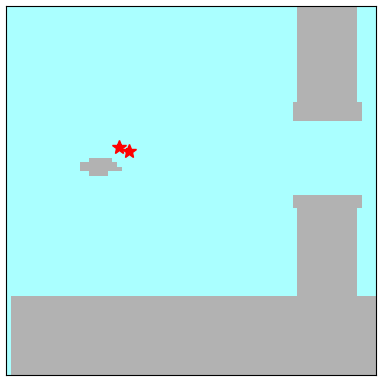}}
        \end{minipage}%
        \par\smallskip        
        \begin{minipage}{.23\textwidth}
        \centering
        \subfloat{\label{fig:flappy-bird-split-group2:a}\includegraphics[scale=.18]{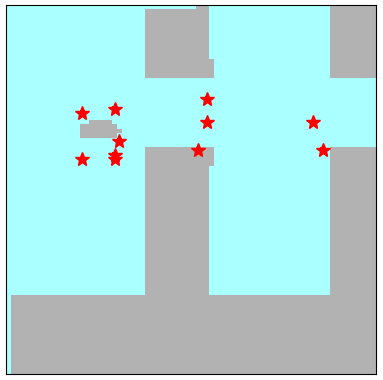}}
        \end{minipage}%
        \begin{minipage}{.23\textwidth}
        \centering
        \subfloat{\label{fig:flappy-bird-split-group2:b}\includegraphics[scale=.18]{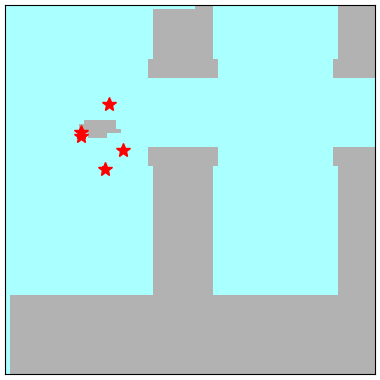}}
        \end{minipage}%
        \begin{minipage}{.23\textwidth}
        \centering
        \subfloat{\label{fig:flappy-bird-split-group2:c}\includegraphics[scale=.18]{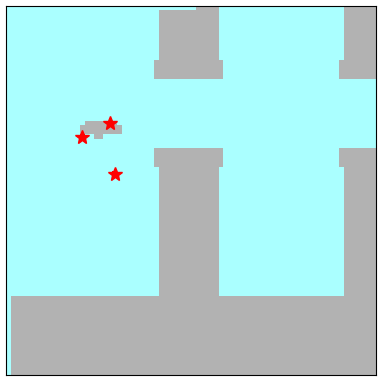}}
        \end{minipage}%
        \begin{minipage}{.23\textwidth}
        \centering
        \subfloat{\label{fig:flappy-bird-split-group2:d}\includegraphics[scale=.18]{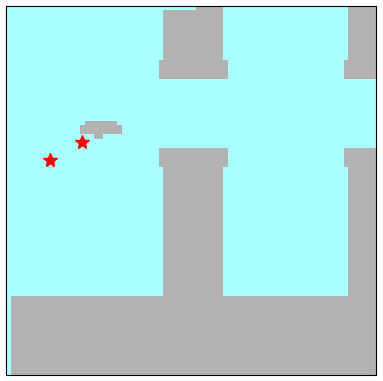}}
        \end{minipage}%
        \caption{Flappy Bird input images with Super-pixels (marked with red star). 
       The input order of four consecutive images is left to right. }
        \label{fig:flappy-bird-split-pixel}
    \end{figure}

    In video games, DRL models take the raw pixels from four consecutive images as input. 
    To mimic the deep models, our LMUT also learns on four continuous images and performs splits directly on raw pixels.
    Deep models for image input can be explained by super-pixels~\cite{ribeiro2016should}. We 
    highlight the pixels that have feature influence $\it{Inf}_{f}>0.008$ (the mean of all feature influences)
    along the splitting path from root to the target partition cell.
    Figure~\ref{fig:flappy-bird-split-pixel} provides two examples of input images with their highlighted pixels at the beginning of game (top) and in the middle of game (bottom).
    We find 1) most splits are made on the first image which reflects the importance of the most recent image input 2) the first image is often used to locate the pipes (obstacles) and the bird, while the remaining three images provide further information about the bird's location and velocity. 

\section{Conclusion}
    This work introduced a mimic learning framework for a Reinforcement Learning Environment. 
    A novel Linear Model U-tree represents an interpretable model with the expressive power to approximate a Q value function learned by a deep neural net. We introduced a novel on-line LMUT mimic learning algorithm based on stochastic gradient descent. 
    %
    Empirical evaluation compared LMUT with five baseline methods on three different Reinforcement Learning environments. The LMUT model achieved clearly the best match to the neural network in terms of its performance on the RL task. 
    We illustrated the abillity of LMUT to extract the knowledge implicit in the neural network model, by (1) computing the influence of features, (2) analyzing the extracted rules and (3) highlighting the super-pixels. 
    A direction for future work is to explore variants of our LMUT, for example by adding tree pruning, and by experimenting more extensively with hyper-parameters. Another important topic is sampling strategies for the active play setting, which would illuminate the difference we observed between matching the neural net's play performance, vs. matching the function it represents. 
\bibliographystyle{splncs04}
\bibliography{bibliography}

\end{document}